\documentclass[10pt,twocolumn,letterpaper]{article}

\usepackage{cvpr}
\usepackage{times}
\usepackage{epsfig}
\usepackage{graphicx}
\usepackage{amsmath}
\usepackage{amssymb,amsbsy,subfigure,amsmath,amsfonts,multirow,color,url,bm,cases,eqparbox}
\usepackage{algorithmic}
\usepackage{algorithm}
\usepackage[mathscr]{eucal}
\usepackage{epstopdf}
\usepackage{epsfig}
\usepackage{float}
\usepackage{makecell}
\usepackage{epstopdf}
\usepackage{graphicx}

\usepackage[breaklinks=true,bookmarks=false]{hyperref}

\cvprfinalcopy 

\def\cvprPaperID{****} 

\begin{document}

\title{Deep Neural Network Based Hyperspectral Pixel Classification With Factorized Spectral-Spatial Feature Representation}

\author{
Jingzhou~Chen\\
Zhejiang University\\
No. 38, ZheDa Road.\\
{\tt\small chengjingzhoucs@zju.edu.cn}
\and
Siyu~Chen\\
Zhejiang University\\
No. 38, ZheDa Road.\\
{\tt\small sychen@zju.edu.cn}\\
\and
Peilin~Zhou\\
Zhejiang University\\
No. 38, ZheDa Road.\\
\and
Yuntao Qian\\
Zhejiang University\\
No. 38, ZheDa Road.\\
{\tt\small ytqian@zju.edu.cn}
}

\maketitle

\abstract{Deep learning has been widely used for hyperspectral pixel classification due to its ability of generating deep feature representation.
However, how to construct an efficient and powerful network suitable for hyperspectral data is still under exploration. In this paper, a novel neural network model is designed for taking full advantage of the spectral-spatial structure of hyperspectral data. Firstly, we extract pixel-based intrinsic features from rich yet redundant spectral bands by a subnetwork with supervised pre-training scheme. Secondly, in order to utilize the local spatial correlation among pixels, we share the previous subnetwork as a spectral feature extractor for each pixel in a patch of image, after which the spectral features of all pixels in a patch are combined and feeded into the subsequent classification subnetwork. Finally, the whole network is further fine-tuned to improve its classification performance. Specially, the spectral-spatial factorization scheme is applied in our model architecture, making the network size and the number of parameters great less than the existing spectral-spatial deep networks for hyperspectral image classification. Experiments on the hyperspectral data sets show that, compared with some state-of-art deep learning methods, our method achieves better classification results while having smaller network size and less parameters.}

\textbf{Keywords:} Hyperspectral pixel classification, deep neural networks, spectral-spatial feature factorization

\section{Introduction}
Hyperspectral imaging has opened up new opportunities for analyzing a variety of materials in remote sensing as it provides rich information on spectral and spatial distributions of distinct materials. One of its most important applications is pixel classification, which is widely applied in material recognition, target detection, geoindexing, and so on~\cite{li2013generalized,huang2008adaptive,du2014discriminative,li2015combined}. However, the classification of hyperspectral image (HSI) still faces some challenges such as, the unbalance between a small number of available training samples and a large number of narrow spectral bands, the high variations of the spectral signature from identical material, high similarities of spectral signatures between some different materials, and the noise impact from the sensors and environment~\cite{bioucas2013hyperspectral}. In order to address the above problems, it is necessary to extract robust and discriminant features. The popular spectral feature extraction algorithms include principal component analysis~(PCA)~\cite{bajorski2011statistical}, independent component analysis~(ICA)~\cite{villa2011hyperspectral}, linear discriminant analysis~(LDA)~\cite{bandos2009classification}, manifold learning~\cite{bachmann2006improved,camps2005kernel}, and various band selection methods~\cite{mendenhall2008relevance,martinez2007clustering,feng2014hyperspectral}. In addition, many studies have demonstrated that it is difficult to well distinguish pixels with spectral information alone, hence the spatial-spectral feature extraction attracts more and more attention. A number of joint spectral-spatial features have been proposed such as extended morphological profiles~\cite{benediktsson2005classification} and 3-D discrete wavelet transform\cite{qian2013hyperspectral}.

Recently, with the great development of modern neural network technique known as deep learning, it has exhibited more beneficial advantages and obtained enormous success in many fields including image segmentation \cite{rota2017loss}, image classification \cite{krizhevsky2012imagenet,sermanet2012convolutional,he2016deep}, artistic style transfer \cite{radford2016unsupervised}, object detection \cite{girshick2014rich}, face verification/identification \cite{sun2014deep,sun2014deep2}, speech recognition \cite{sainath2013deep} and translation \cite{vaswani2017attention}. Comparing to classic statistical methods that explicitly designate fully specified modeling procedures, deep learning tries to fit an implicit yet potentially powerful function that can both imitate bionic mechanism and extract sophisticated features by a data-driven learning process. A number of state-of-art deep learning methods have been applied to HSI processing, for example, unmixing \cite{ozkan2018deep}, target detection \cite{pan2016detection}, HSI visualization \cite{Chen2018Spectral}, HSI denoising \cite{xie2017hyperspectral}, HSI super-resolution \cite{LI201729} and HSI classification \cite{chen2014deep,windrim2018pretraining,hu2015deep,ma2016spectral,mei2017learning,romero2016unsupervised,8061020,8283837,li2017spectral,chen2016deep,li2017hyperspectral,liu2018supervised}.

Among them, the deep neural network based approaches for HSI classification can also be sketchily categorized into spectral based methods and spectral-spatial based methods. The first type of deep learning methods directly use the spectral signatures of pixels for classification, including stacked autoencoder based methods \cite{chen2014deep}, pre-training based methods \cite{windrim2018pretraining}, and convolutional neural network (CNN) based method \cite{hu2015deep}. Even though  they bring notable improvement of HSI classification over conventional classification techniques such as k-nearest-neighbor (KNN) and support vector machine (SVM), some pixels are still difficult to be accurately classified using the spectral information alone due to the high inter-class similarity and the high intra-class difference. On the other hand, spectral-spatial information based deep-learning approaches have been proved to achieve better performance than those spectral information based ones. The examples are spatial stacked autoencoder (SAE) based method \cite{ma2016spectral}, CNN based spectral-spatial feature extraction methods \cite{mei2017learning,romero2016unsupervised} , spectral-spatial  CNN based classifiers \cite{li2017spectral,8061020,8283837}. Particularly, three dimensional (3D) CNN \cite{chen2016deep} is a typical model used in most of these spectral-spatial information based deep-learning approaches. 3D CNN modifies standard CNN to convolve along both spatial and spectral dimensions for HSI classification. Such a scheme can always employ local spatial information in HSI patches to perform classification learning and inference. However, compared with the  1D and 2D CNN, 3D CNN requires a greater size of model to capture useful features as the number and size of kernels grow rapidly in respect to input size and dimension, especially when the spatial dimension and spectral dimension are distinct in terms of physical mechanism. In general, 3D CNN tends to have excessive parameters, so as to be prone to over-fit and hard to train. As has been suggested by K. He \emph{et al.} in \cite{he2016deep}, the oversized network may encounter certain approximation difficulties. Han \emph{et al.} also pointed out in \cite{han2015deep} that the weights of CNN may be redundant and most of them do not carry significant information. Li \emph{et al.} in \cite{li2017bt} showed a similar fact that prevailing deep models suffer from weight redundancy problem. Zhang \emph{et al.} in \cite{zhang2017understanding} also indicated that the oversized models with the excessive amount of parameters often tend to memorize data sets instead of learning the generic task solutions. Furthermore, on account of the limited resources of class-labeled pixels in hyperspectral datasets, this problem will cause severer degradation of classification performance. Therefore, reducing the size of model becomes a significant problem for spectral-spatial CNN based HSI pixel classification methods.

Besides of the routine techniques of model compression for deep learning models such as normalization, regularization, and  network pruning,
operation factorization is popularly used in CNN and other deep neural networks to decouple a complex computation into many much smaller steps which have far less parameters in total \cite{szegedy2015going,hu2017squeeze}. Such as in \cite{howard2017mobilenets}, a convolutional layer with $73728$ parameters are factorized into two separate operations: one depth-wise and one point-wise convolution, there are totally $576+8192=8768$ parameters. In \cite{szegedy2016rethinking}, $n\times n$ convolution is factorized into a $1\times n$ convolution followed by a $n\times 1$ convolution. The decoupled/factorized operations can improve the generalization capability, consume less resources, and make training and inference faster. Therefore, operation factorization is also one of the prevailing as well as practical schemes applied in spectral-spatial deep learning models for HSI classification. For example, in \cite{li2017hyperspectral}, CNN with pixel-pair features (CNN PPF) is based on the similarity of local constituents, which takes a pair of pixels as input, and the output tells if these two pixels are of different classes or which class they all belong to. The spectral-spatial feature in CNN-PPF is factorized into class related latent spectral feature and pixel-pair consistency based spatial feature. The final label of the target pixel is determined via a voting strategy based on the neighboring pixel-pair information. CNN-PPF performs well using its augmented data and $L_2$ regularization scheme. Similarly, in \cite{liu2018supervised},  Siamese CNN (S-CNN) uses paired image patches as training samples, and trains them to classify the central pixels of patches, in which the spectral-spatial feature is factorized into the spectral feature extracted by Siamese network and the spatial information used by an SVM that combines the spectral feature vectors of patches outputted by Siamese network. Unfortunately, the combination schemes of the spectral and spatial operations in both of CNN-PPF  and S-CNN are fixed rather than adaptively learned, as the voting strategy is used by CNN-PPF, and a separated SVM is used in S-CNN. Therefore an end-to-end CNN with global optimization cannot be achieved.

\begin{figure}[tbp]
  \centering
  \includegraphics[width=\linewidth]{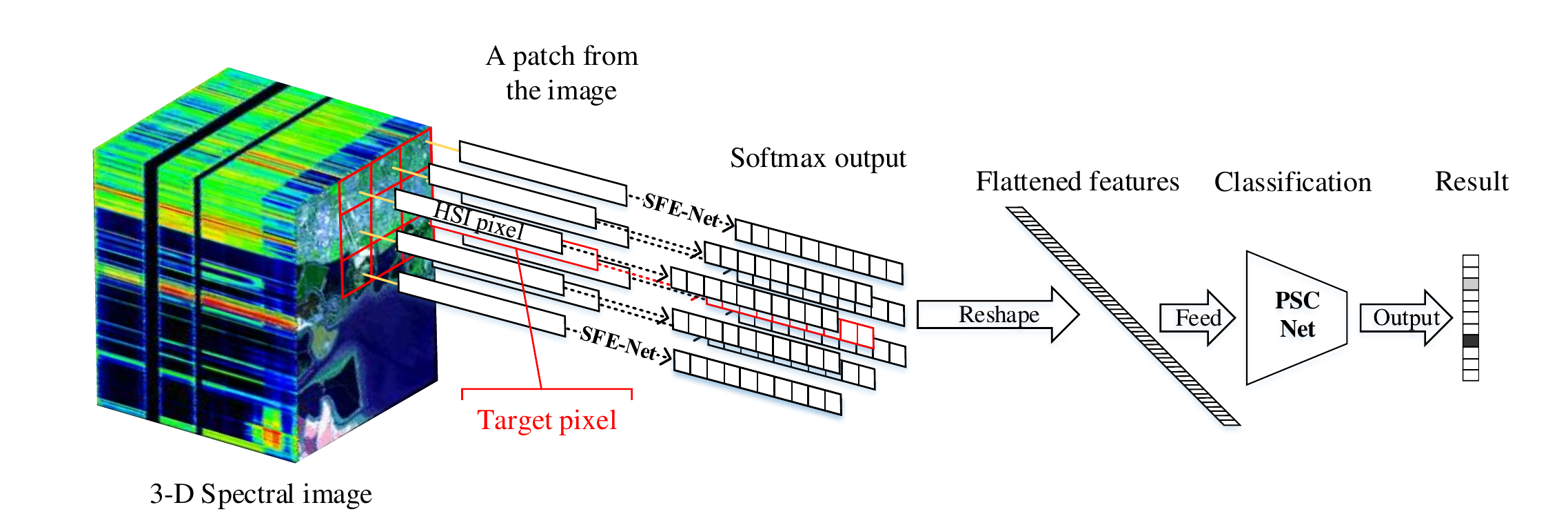}
  \caption{Detailed workflow of FSSF-Net for predicting the target pixel}\label{fig:detailed_workflow}
\end{figure}

In this paper, we propose a deep neural network that utilizes both spectral and spatial information in a novel decoupled/factorized manner, which is designed to be concise, adaptive, end-to-end and easy to train. Our spectral-spatial classification model is factorized into a pixel-based spectral feature extraction subnetwork (SFE-Net) and a patch based spatial classification subnetwork (PSC-Net). The SFE-Net is pre-trained in a supervised learning scheme, as a rough prediction based on spectral information alone, and it can be constructed like any spectrum-based deep neural network. In order to refine our prediction, the PSC-Net is then trained to determine the class of the central pixel in a patch by taking the prior spectral information of surrounding pixels into account, as there are strong spatial relationships among these neighboring pixels. These two subnetworks are combined via a sequential connection. We first share the SFE-Net across each pixel in the patch, then concatenate all the extracted spectral features as the input vector of the PSC-Net. Besides, two factorized subnetworks are trained collaboratively as an entire network by the back-propagation algorithm to improve the overall performance. Because both of SFE-Net and PSC-Net used in the paper are based on deep neural networks, and spectral-spatial features are factorially represented, we name the proposed model \textit{Factorized Spectral-Spatial Feature Network} (FSSF-Net). A demonstration of the proposed framework is shown in Fig. \ref{fig:detailed_workflow}.

In summary,  a novel deep learning based framework is proposed in the paper that better utilizes joint spectral-spatial structure of HSI  by a flexible and generalized feature factorization scheme. It allows various kinds of deep neural networks to act as subnetworks or hidden layers, meanwhile it is an integrated end-to-end model. Compared to some state-of-the-art deep learning based HSI classification methods including CNN based ones, the proposed approach achieves competitive classification performance, especially the light-weight model resulted from feature factorization brings faster training, lower storage, and better generalization with a small number of class-labeled samples.

The paper is organized as follows. In Section \ref{sec:proposed_method}, we introduce the mathematical formulation of the problem and the proposed model, followed by the details of the implementation. In Section \ref{sec:experiments}, a number of experiments are done to investigate the effectiveness of the proposed model and the primary techniques used in it, and experimental comparisons with some state-of-the-art approaches are also given. Finally, we conclude our work in Section \ref{sec:conclusion}.



\section{Methods}\label{sec:proposed_method}

In this section, we further explain some details of our network including specific network architecture and training scheme. We first highlight the detail of two subnetworks, SFE-Net and PSC-Net, and how they are connected together. As one of the major concern, training scheme is also well explained, especially how the error is back propagated through PSC-Net to SFE-Net.

\subsection{Model Architecture}\label{sec:implementation}


%
%

FSSF-Net takes a 3D patch extracted from an HSI as input. As aforementioned, there are two parts in FSSF-Net: SFE-Net and PSC-Net. Given a 3D patch, we first apply and share the SFE-Net across each pixel to extract spectral features. Concatenating these spectral features into a vector to fuse spatial information embodied in the patch. Utilizing this vector, PSC-Net infers the class label of the central pixel.

\begin{figure*}[t]
  \centering
  \subfigure[Architecture of SFE-Net]{
  \includegraphics[width=0.7\linewidth]{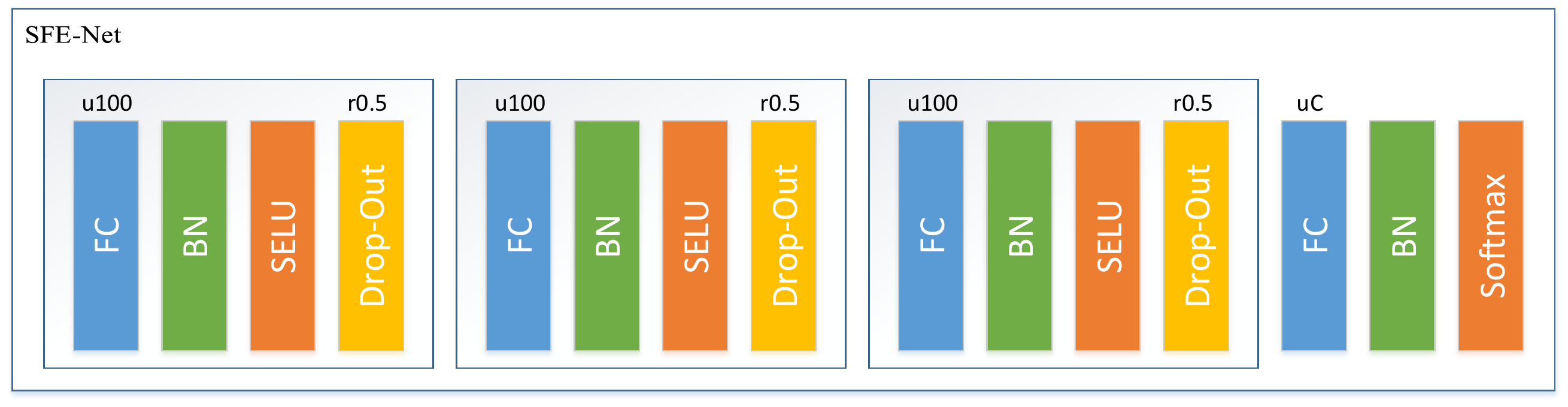}
  }
  \subfigure[Architecture of PSC-Net]{
  \includegraphics[width=0.23\linewidth]{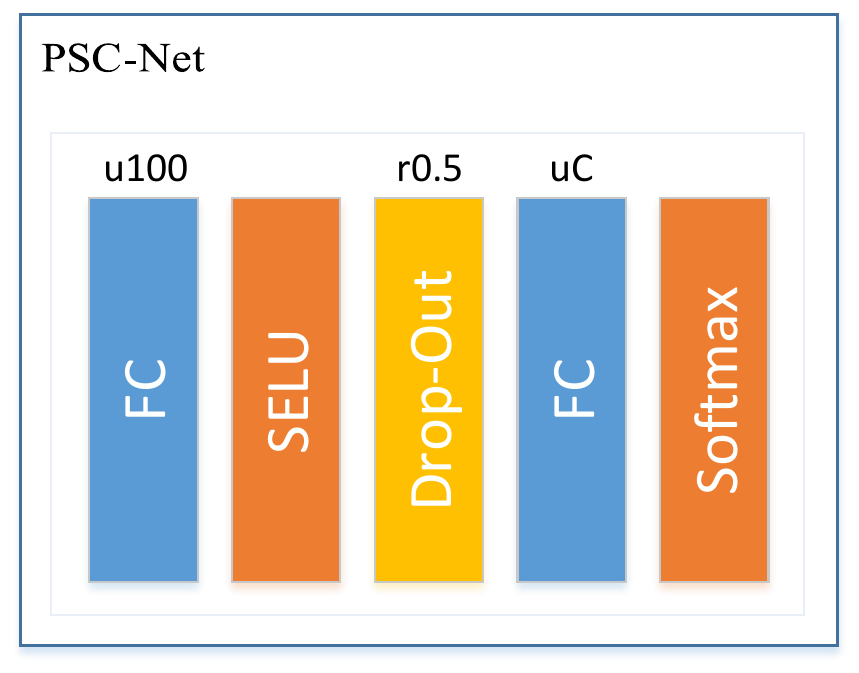}
  }
  \caption{Architectures of  SFE-Net and PSC-Net.}
  \label{fig:architecture}
\end{figure*}

Generally, there is no restriction on the structure and the number of layers in these two subnetworks. SFE-Net and PSC-Net are assumed to be any type of neural networks, i.e., they are not limited to the specific form. For example, the classic 1D-CNN and SAE can be used. In the case of the classification task on hand, the flexibility of our framework allows us to choose and/or develop any network structure in order to maximize the performance. In this paper, we use MLP (Multilayer Perceptron) based architectures for both subnetworks. To improve the generalization capability and advert over-fitting, batch-normalization layers~(BN), self-normalizing ELU~(SELU) layers and Drop-out layers~(Drop-out) are added to the architecture as regularization/normalization components. The network architecture for SFE-Net and PSC-Net is illustrated in \ref{fig:architecture}, where FC stands for fully-connected layers. For both SFE-Net and PSC-Net, the hidden fully-connected layers have the same amount of output units $u=100$ and all their Drop-out layers share the same probability of retaining $r=0.5$. Both SFE-Net and PSC-Net output $C$-dimensional vector activated by softmax as the final output, where $C$ is the number of classes.




\subsection{Model Training}

An patch based HSI dataset can be represented as $X = \{x_i|x_i \in \mathbb{R}^{W\times W \times D},i=1,2,\dots,N\}$, where $W$ is the width of the square HSI patch, $D$ is the the number of spectral bands/channels, $N$ is the number of patches in the data set. The corresponding classification label set is $Y =\{y_i| y_i \in \mathbb{Z}^{C}, i=1,2,\dots,N \}$ where each element $y_i$ is the one-hot label of the central pixel in $x_i$, and $C$ still is the number of classes. The pixel classification of HSI is to find a function $f(x)$ such that the output $Y'=\{y'_i | y'_i = f(x_i),i=1,2,...,N\}$ is as close to classification label $Y$ as possible.

In FSSF-Net, the submodule SFE-Net is pre-trained using classification supervision, so as to provide softmax vectors as output spectral features. Then, the submodule PSC-Net takes these spectral features of an HSI patch and outputs the final classification result, which is also achieved by using a classification loss metric. In summary, both SFE-Net and PSC-Net are trained using \textit{cross-entropy} loss metric. Specifically, given an output vector $y'$ and its corresponding one-hot label $y$, with the j-th element in vector $y$ indexed by $y^{(j)}$, the loss function can be formulated as
\begin{equation}\label{eq:loss_function}
  \mathcal{L} = \frac{1}{N} \sum_{i=1}^{N} \sum_{j=1}^{C} y_i^{(j)} \log(y'^{(j)}_i).
\end{equation}

The training of FSSF-Net includes two stages: pre-training SFE-Net, and fine-tuning SFE-Net, PSC-Net together. In the pre-training stage, we extract the central pixels in each patch $x_i$ and train SFE-Net with corresponding labels. As for the fine-tuning stage, FSSF-Net uses 3D patches $x_i$ as input, whose label is provided by the central pixel. During this stage, both SFE-Net and PCS-Net are set to be trainable. The entire training process completes when the fine-tuning training of FSSF-Net converges. Especially, when back-propagating the error through PSC-Net to SFE-net, the error used for updating SFE-Net is the average of errors calculated on each pixel in a patch.
ADAM (Adaptive Moment Estimation \cite{kingma2015adam:}) optimizer is used in our experiments with a learning rate set to $0.001$ in both stages. The input patch size $W$ is set to 7, which is a tradeoff between the workload of training storage and the amount of spatial information fed into the network. Overall, there are four hyper-parameters $C,W,u,r$ in our implementation where $W,u,r$ are constant through all our experiments.
\section{Results and Discussion}\label{sec:experiments}

In this section, we first introduce five data sets used in our experiments and clarify corresponding experimental settings. Then, we examine three significant ideas applied in our method: the pre-training of SFE-Net ($g$), the sharing of such spectrum extractor spatially, and the flexibility of types of hidden layers. Finally evaluate our method further by comparing with some state-of-art approaches. Tensorflow based on CUDA library is selected as the computational framework, and the unified interface wrapper Keras is applied to simplify implementation development.
They are running on a workstation equipped with an Intel Xeon E5-2620 v4 with 2.1 GHz and Nvidia GeForce GTX 1080 graphics card.
Overall accuracy (OA), average accuracy (AA) and kappa coefficient are used as the criteria of classification accuracy. All experiments were repeated five times.

\subsection[Experimental Data]{Data for Experiments\footnote{\url{http://www.ehu.eus/ccwintco/index.php/Hyperspectral_Remote_Sensing_Scenes}}}

\begin{figure}[htp]
  \centering
  \includegraphics[width=0.2\linewidth]{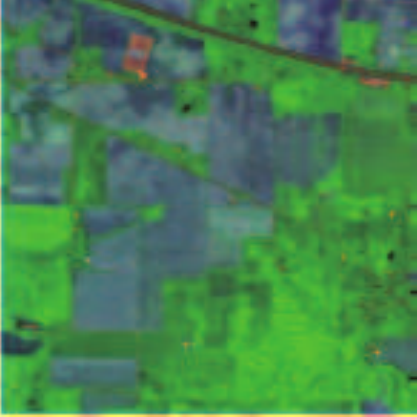}
  \caption{Pseudocolor image (Bands 5, 50, 200) of Indian Pines}\label{fig:false_IN}
\end{figure}

The first is Indian Pines data set gathered by Airborne Visible/Infrared Imaging Spectrometer (AVIRIS) in 1992 from Northwest Indiana, including 16 vegetation classes. There are 220 spectral bands in the 0.4-45 $\mu$m region of the visible and infrared spectrum and have $145 \times 145$ pixels for each band. The pseudocolor image of the Indian Pines data set is shown in Fig.~\ref{fig:false_IN}.

\begin{figure}[htp]
  \centering
  \includegraphics[angle=90]{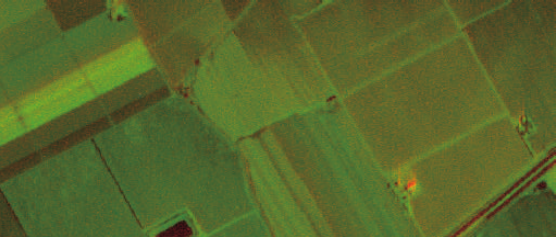}
  \caption{Pseudocolor image (Bands 1, 102, 204) of Salinas}\label{fig:false_SA}
\end{figure}

The second data set Salinas is also collected by the AVIRIS sensor over Salinas Valley, California, including $512 \times 217$ pixels for each band. After 20 water absorption bands are removed, 204 bands remain for the experiments. There are 16 classes contained in the image such as vegetables, bare soils, and vineyard fields. Its pseudocolor image is illustrated in Fig.~\ref{fig:false_SA}.

The Third is KSC dataset still acquired by the AVIRIS sensor over the Kennedy Space Center (KSC), Florida, in 1996. After removing water absorption and low SNR bands, 176 bands were used. There are 13 land-cover classes and $512 \times 614$ pixels with 5211 labeled. Its pseudocolor image is shown in Fig.~\ref{fig:false_SA}.

\begin{figure}[htp]
  \centering
  \includegraphics[width=0.2\linewidth,angle=90]{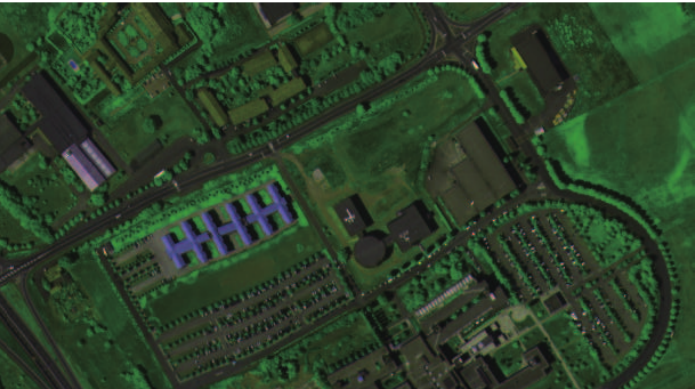}
  \caption{Pseudocolor image (Bands 10, 50, 80) of Pavia University}\label{fig:false_PU}
\end{figure}

The fourth is the University of Pavia data set acquired by Reflective Optics System Imaging Spectrometer (ROSIS) in Northern Italy in 2001.
The image scene contains 9 urban land-cover types and $610 \times 340$ pixels for each band. Once the very noisy bands have been removed, the remaining 103 spectral bands, in the 0.43-0.86 $\mu$m range of the visible and infrared spectrum, are employed. Its pseudocolor image is shown in Fig.~\ref{fig:false_PU}.

\begin{figure}[htp]
  \centering
  \includegraphics[width=0.2\linewidth,angle=90]{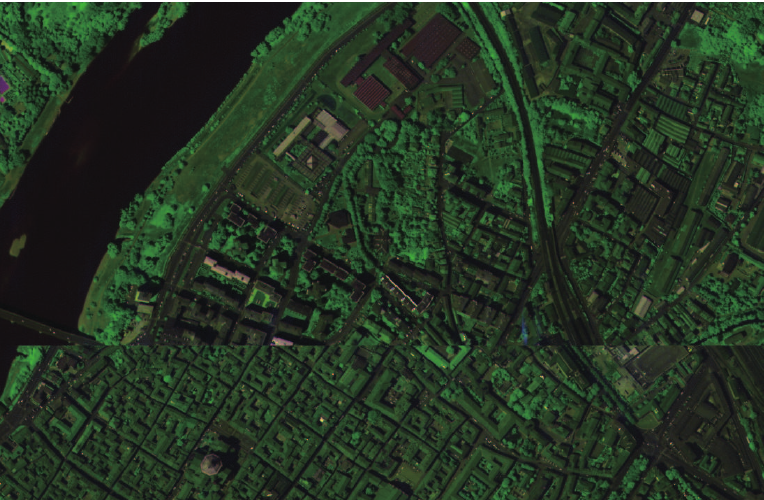}
  \caption{Pseudocolor image (Bands 1, 30, 80) of Pavia Center}\label{fig:false_PA}
\end{figure}

The fifth is the Pavia Center data set also acquired by the ROSIS sensor over Pavia, northern Italy. After removing noise bands, there are 102 spectral bands left. The original size of Pavia Center is $1096 \times 1096$, but some of the samples in the image contain no information and have to be discarded, thus a subset with a size of $1096 \times 715$ is selected, which includes 9 different classes.  Its pseudocolor image can be seen in Fig.~\ref{fig:false_PA}.

\begin{figure}[htp]
  \centering
  \includegraphics[width=0.25\linewidth]{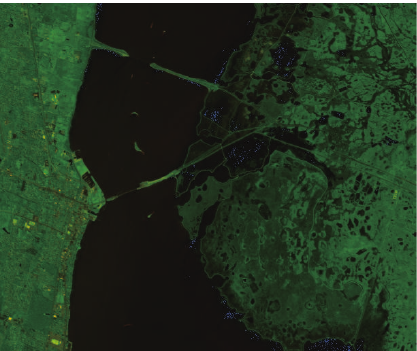}
  \caption{Pseudocolor image (Bands 5, 50, 150) of KSC}\label{fig:false_KSC}
\end{figure}

\subsection{Model Setting}\label{sec:framework_setting}

We implement the proposed architecture, as stated in Section.~\ref{sec:proposed_method}. Specially, considering our network adapted to different data sets, the spectral dimension may differ in the input layer of SFE-Net. The number of neurons in the last softmax layer of both SFE-Net and PSC-Net may also vary corresponding to the number of classes~($C$) in each data set.

Besides, optimization process is also a very important factor influencing the classification performance.
We adopt the ADAM~\cite{kingma2015adam:} optimizer to harness and speed up the training process.  The initial learning rate is set to 0.001 for both pre-training and fine-tuning.
The learning rate of decay is set to 0.005 for the pre-training, while set to 0.01 for the fine-tuning.
Considering the limited training data in our task, we are able to set the batch size to be equal to the whole training set for better gradient behaviour.
The pre-training and fine-tuning have 10000 epochs and 1000 epochs respectively.

Last but not least, it is worth noting that we fix all hyperparameters mentioned above in all experiments, which means that we do not tune our model to overfit any specific data sets.

\subsection{Effectiveness of Proposed Model}

Here we will evaluate the effectiveness of three significant ideas applied in our proposed architecture, which are the supervised pre-training of SFE-Net for extracting spectral feature, the sharing of such feature extractor spatially to include the local spatial correlation into PSC-Net, and the flexibility of network type for hidden layers.
In addition, during the fine-tuning stage, the whole FSSF-Net is trained, including the pre-trained network SFE-Net. Therefore, we demonstrate whether these three schemes in the design of the architecture of our model actually contribute to improving accuracy in the following experiments.

\subsubsection{Supervised pre-training}\label{sec:supervised_pre_training}

To verify the necessity of pre-training, we compare the model that the parameters in SFE-Net $g$ are initialized by the pre-training process with the model whose parameters of $g$ are randomly initialized. In other words, the structure of the whole model FSSF-Net remains the same, and only the parameter initializations in the network $g$ differ.

The University of Pavia and the Indian Pines data sets are chosen to perform such comparison, and the training sets are selected with the small and medium sizes.
For the Indian Pines data set, we randomly selected $5\%$ and $10\%$ of the labeled samples from each class to form the training sets respectively, leaving the rest as the test sets. As for the Pavia University data set, because the training samples have been separated from the total labeled ones, see Table~\ref{fixed_training_PU}, we randomly selected $1\%$ and $10\%$ samples from each class in the separated training samples as the training sets, and all the remaining labeled samples are used for test sets. The compared results are displayed in Table~\ref{pre-train}, from which we can observe that both in the Pavia University and in the Indian Pines, the supervised pre-training can significantly improve the classification accuracy, especially in the case of small training-sample size.

\begin{table}[htp]
  \centering
  \caption{SEPARATION OF TRAINING SAMPLES FOR PAVIA UNIVERSITY DATA SET}\label{fixed_training_PU}
  \begin{tabular}{cccc}
    \hline
    \# & Class & Samples & Training Samples \\
    \hline
    1 & Asphalt & 6631 & 548 \\
    2 & Meadows & 18649 & 540 \\
    3 & Gravel & 2099 & 392 \\
    4 & Trees & 3064 & 524 \\
    5 & Painted metal sheets & 1345 & 265 \\
    6 & Bare Soil & 5029 & 532 \\
    7 & Bitumen & 1330 & 375 \\
    8 & Self-Blocking Bricks & 3682 & 514 \\
    9 & Shadows & 947 & 231 \\
    \hline
  \end{tabular}
\end{table}

\begin{table}[htp]
  \centering
  \caption{THE CLASSIFICATION RESULTS WITH AND WITHOUT PRE-TRAINING~(\%)}\label{pre-train}
  \begin{tabular}{|c|c|c|c|c|c|}
    \hline
    \multicolumn{2}{|c}{Dataset} & \multicolumn{2}{|c}{Pavia University} & \multicolumn{2}{|c|}{Indian Pines} \\
    \hline
    Pretraining & Metrics & 1\% & 10\% & 5\% & 10\% \\
    \hline
    \multirow{3}*{Yes} & OA & \textbf{73.51} & \textbf{97.21} & \textbf{95.03} & \textbf{98.08} \\\cline{2-6}
      & AA & \textbf{84.07} & \textbf{97.69} & \textbf{82.64} & \textbf{93.86} \\\cline{2-6}
      & Kappa & \textbf{67.54} & \textbf{96.32} & \textbf{94.33} & \textbf{97.81} \\\cline{2-6}
    \hline
    \multirow{3}*{No} & OA & 68.21 & 96.06 & 89.92 & 95.47 \\\cline{2-6}
      & AA & 74.21 & 96.57 & 75.58 & 88.24 \\\cline{2-6}
      & Kappa & 60.32 & 94.80 & 88.47 & 94.84 \\\cline{2-6}
    \hline
  \end{tabular}
\end{table}

\subsubsection{Sharing of pre-trained network}

 Besides of the pre-training scheme, parameter sharing is also a popular strategy embodied in a variety of deep learning models. For example, CNN extracts shifting invariant features from images by the convolutional operation, and recurrent neural network (RNN) models the sequential dependence by designing the network with loops in them.
 Different from these sharing methods implied in the scenario of CNN and RNN, we assume that the adjacent pixels around a central pixel share similar spectral information, therefore it is natural to use the same pre-training network $g$ to extract spectral features from each pixel belonging to a $7 \times 7 \times D$ patch.
 To verify the effectiveness of this sharing architecture, we set up a contrastive experiment. First, we use the same network $g$ to extract each pixel's spectral features in a patch. Second, on the contrary, for different pixels in a patch, different networks are used, which have the same structure as the network $g$ but do not share their parameters.
 In the later case, the parameters of those networks are still initialized from the pre-training network, but during the fine-tuning process, those networks for different pixels in a patch may learn different parameters.

We apply the same experimental setting in Section~\ref{sec:supervised_pre_training} and classification results are displayed in Table~\ref{sharing_architecture}. Seen from Table~\ref{sharing_architecture}, classification accuracy has been dramatically boosted due to applying the sharing scheme.
The improvement with sharing scheme in the Indian Pines data set is more obvious than that in the Pavia University data set, because the local spatial correlation is stronger in the Indian Pines data set than in the Pavia University data set.
In addition, with the less number of parameters of the whole network that adopts the sharing scheme, it is less likely to overfit as the model complexity is reduced.

\begin{table}[htp]
  \centering
  \caption{THE CLASSIFICATION RESULTS WITH AND WITHOUT SHARING SCHEME~(\%)}\label{sharing_architecture}
  \begin{tabular}{|c|c|c|c|c|c|}
    \hline
    \multicolumn{2}{|c}{Dataset} & \multicolumn{2}{|c}{Pavia University} & \multicolumn{2}{|c|}{Indian Pines} \\
    \hline
    Param-Sharing & Metrics & 1\% & 10\% & 5\% & 10\% \\
    \hline
    \multirow{3}*{Yes} & OA & \textbf{73.51} & \textbf{97.21} & \textbf{95.03} & \textbf{98.08} \\\cline{2-6}
      & AA & \textbf{84.07} & \textbf{97.69} & \textbf{82.64} & \textbf{93.86} \\\cline{2-6}
      & Kappa & \textbf{67.54} & \textbf{96.32} & \textbf{94.33} & \textbf{97.81} \\\cline{2-6}
    \hline
    \multirow{3}*{No} & OA & 72.57 & 93.86 & 80.40 & 86.51 \\\cline{2-6}
      & AA & 74.53 & 94.39 & 67.74 & 77.40 \\\cline{2-6}
      & Kappa & 64.54 & 91.89 & 77.47 & 84.53 \\\cline{2-6}
    \hline
  \end{tabular}
\end{table}

\subsubsection{Flexibility of network type in hidden layers}

When applying the proposed framework to build our classification network, we adopt MLP, or called fully connected network (FCN)  as hidden layers. Specifically, the hidden layers in SFE-Net $g$ are FCN together with dropout and batch normalization layers, see Fig.~\ref{fig:architecture}(a),
and the hidden layers in PSC-Net are FCN with dropout layer, see Fig.~\ref{fig:architecture}(b)).
However, the main contribution of this paper is proposing an network architecture suitable to fuse the spatial-spectral information in HSI,
while whichever network type used to implement this architecture is flexible.
To explore the flexibility of the proposed architecture, we compare the FCN with locally connected network (LCN) and convolutional neural network (CNN), because CNN and LCN are also widely used for hidden layers as FCN.
In fact, FCN layer in network $g$ can be viewed as a convolutional layer with kernel size $(1,1,D)$ and stride size $(1,1,0)$, and the number of convolutional kernels equals the number of neurons in the FCN layer. On the contrast, LCN is also considered as a CNN but it does not share the parameters in the convolutional process.
In the comparative experiment, we only replace the first two hidden layers in Fig.~\ref{fig:architecture}(a) with LCN or CNN layers, keeping rest structure in the whole network the same as the case of FCN. In the case of LCN, the first hidden layer in network $g$ contains 20 convolutional kernels with kernel size $(1,1,5)$ and stride size $(1,1,3)$, and the second hidden layer includes 15 output channels with the same kernel and stride sizes as the first layer.
As for CNN, to make a fair comparison, the first two hidden layers have the same number of kernels and kernel structure as LCN.

The Pavia University dataset was used in the experiment, 50 samples were randomly selected from each class as training samples and the compared results are displayed in
Table~\ref{fig:flexibility}, in which spectral only model is just SFE-Net used for classification, and spectral-spatial model is the whole network including SFE-Net and PSC-Net.
We can find when taking adjacent spatial information into consideration, the classification performance can be  further improved whichever type of hidden layers is used.
Moreover, LCN and CNN receive competitive results compared with FCN when spatial-spectral information is used.
These facts indicate that the classification improvement comes from the proposed architecture rather than any specific type of hidden layers.

\begin{table}[htb]
  \centering
  \caption{THE CLASSIFICATION RESULTS WITH DIFFERENT TYPES OF HIDDEN LAYERS  IN SFE-NET~(\%)}\label{fig:flexibility}
  \begin{tabular}{|c|c|c|c|c|}
    \hline
    Training & Metrics & FCN  & LCN & CNN \\
    \hline
    \multirow{3}*{Spectral only} & OA & \textbf{81.37} & 76.14 & 76.09 \\\cline{2-5}
                                & AA & \textbf{85.40} & 84.91 & 84.05 \\\cline{2-5}
                                & Kappa & \textbf{75.99} & 70.16 & 69.82 \\\cline{2-5}
    \hline
    \multirow{3}*{Spectral-spatial} & OA & \textbf{96.97} & 96.43 & 96.72 \\\cline{2-5}
                                    & AA & 97.77 & 97.57 & \textbf{98.04}\\\cline{2-5}
                                    & Kappa & \textbf{96.00} & 95.30 & 95.68 \\\cline{2-5}
    \hline
  \end{tabular}
\end{table}

\subsection{Comparison with Other Methods}

To further demonstrate the effectiveness of the proposed method, we compare it with some traditional prominent methods such as spectral feature based SVM~\cite{waske2010sensitivity}, 3D-DWT~\cite{qian2013hyperspectral} spectral-spatial feature based SVM, and some state-of-art deep learning methods, like CNN~\cite{hu2015deep},
3D-CNN~\cite{chen2016deep}, CNN-PPF~\cite{li2017hyperspectral}, S-CNN~\cite{liu2018supervised}.
In the experiments, SVM has radial basis function~(RBF) kernel, and is implemented by the \emph{libsvm} toolbox.
As described earlier, our proposed model is linked to the convolutional operation, therefore, we compare our method with several CNN based methods.
CNN applies the convolutional operation into hyperspectral image classification but it only considers the spectral information.
3D-CNN inputs a 3D patch and convolves it with 3D convolutional kernels.
CNN-PPF and S-CNN adopt data augmentation by paired similar pixels.

It is difficult to make a fair comparison between various deep learning methods because there are many hyperparameters needed to be adjusted properly.
Therefore, firstly we compare our method to each one of those deep learning methods (CNN, 3D-CNN, CNN-PPF, S-CNN) under the same setting as they are stated in their original papers respectively, and the classification results of these four methods are directly copied from their papers since the reported results are obtained with the optimal or near optimal structures and parameters tuned by those authors.
According to the papers of CNN and CNN-PPF, Salinas, Indian Pines, and Pavia University are used as experimental data sets, and we denote their experimental seting as "setting 1" where 200 samples are randomly selected from each class as the training samples and leave the rest as the test samples.
The "setting 2" refers to the experimental settings in the paper of 3D-CNN, in which
the separations of training and testing samples on Kennedy Space Center, Indian Pines, and Pavia University are illustrated in Tables~\ref{3d_cnn_IN}, \ref{3d_cnn_KSC}, \ref{3d_cnn_PU} respectively.
The "setting 3" is based on the paper of S-CNN, in which 200 samples are randomly selected from each class as training samples in Pavia Center, Indian Pines, and Pavia University
data sets respectively, whereas the whole labeled samples in each data set as test samples.

For all three "settings", we obtain the results of the proposed FSSF-Net method with the before mentioned architecture and parameters in Section \ref{sec:framework_setting}.
The comparing results of all five deep learning methods in three "settings" are gathered in Table~\ref{various_settings}.
In addition to classification accuracy, we also evaluate the complexities of these models by calculating their numbers of parameters for different data sets, which are recorded in Table~\ref{various_parameters}.
Overall speaking, observed from Table~\ref{various_settings} and Table~\ref{various_parameters}, our method reaches a good tradeoff between the model complexity and the classification accuracy. CNN and CNN-PPF have fewer parameters than ours, but 3D-CNN and S-CNN contain much more parameters than ours.  Our method obtains better results than those of CNN and CNN-PPF, and outperforms 3D-CNN and S-CNN in most cases.

\begin{table}[htp]
  \centering
  \caption{EXPERIMENTAL SETTING OF INDIAN PINES IN SETTING 2}\label{3d_cnn_IN}
  \footnotesize
  \begin{tabular}{cccc}
    \hline
    Number & Class & Training & Test \\
    \hline
    1 & Alfalfa & 30 & 16 \\
    2 & Corn-notill & 150 & 1198 \\
    3 & Corn-min & 150 & 232 \\
    4 & Corn & 100 & 5 \\
    5 & Grass-pasture & 150 & 139 \\
    6 & Grass-trees & 150 & 580 \\
    7 & Grass-pasture-mowed & 20 & 8 \\
    8 & Hay-windrowed & 150 & 130 \\
    9 & Oats & 15 & 5 \\
    10 & Soybean-notill & 150 & 675 \\
    11 & Soybean-mintill & 150 & 2032 \\
    12 & Soybean-clean & 150 & 263 \\
    13 & Wheat & 150 & 55 \\
    14 & Woods & 150 & 793 \\
    15 & Buildings-Grass-Trees & 50 & 49 \\
    16 & Stone-Steel-Towers & 50 & 43 \\
    \hline
      & Total & 1765 & 6223 \\
    \hline
  \end{tabular}
\end{table}

\begin{table}[htp]
  \centering
  \caption{EXPERIMENTAL SETTING OF KSC IN SETTING 2}\label{3d_cnn_KSC}
  \footnotesize
  \begin{tabular}{cccc}
    \hline
    Number & Class & Training & Test \\
    \hline
    1 & Scrub & 33 & 314 \\
    2 & Willow swamp & 23 & 220 \\
    3 & CP hammock & 24 & 232 \\
    4 & Slash pine & 24 & 228 \\
    5 & Oak/Broadleaf & 15 & 146 \\
    6 & Hardwood & 22 & 207 \\
    7 & Swamp & 9 & 96 \\
    8 & Graminoid marsh & 38 & 352 \\
    9 & Spartina marsh & 51 & 469 \\
    10 & Cattail marsh & 39 & 365 \\
    11 & Salt marsh & 41 & 378 \\
    12 & Mud flats & 49 & 454 \\
    13 & Water & 91 & 836 \\
    \hline
      & Total & 459 & 4297 \\
    \hline
  \end{tabular}
\end{table}

\begin{table}[htp]
  \centering
  \caption{EXPERIMENTAL SETTING OF PAVIA UNIVERSITY IN SETTING 2}\label{3d_cnn_PU}
  \footnotesize
  \begin{tabular}{cccc}
    \hline
    Number & Class & Training & Test \\
    \hline
    1 & Asphalt & 548 & 5472 \\
    2 & Meadows & 540 & 13750 \\
    3 & Gravel & 392 & 1331 \\
    4 & Trees & 542 & 2573 \\
    5 & Metal Sheets & 256 & 1122 \\
    6 & Bare soil & 532 & 4572 \\
    7 & Bitumen & 375 & 981 \\
    8 & Bricks & 514 & 3363 \\
    9 & Shadows & 231 & 776 \\
    \hline
      & Total & 3930 & 33940 \\
    \hline
  \end{tabular}
\end{table}

\begin{table*}[htp]\footnotesize
  \centering
  \caption{CLASSIFICATION RESULTS (\%) UNDER DEFFIRENT EXPETIMENTAL SETTINGS}\label{various_settings}
  \begin{tabular}{|c|c|c|c|c|c|c|c|}
    \hline
    Experimental Setting & Data Set & Metrics & CNN & 3D-CNN & CNN-PPF & S-CNN & FSSF-Net \\
    \hline
    \multirow{9}*{Setting 1} & \multirow{3}*{Salinas} & OA & 92.60 & & 94.80 & & \textbf{95.98} \\
                              &                    & AA & & & & & 98.68 \\
                              &                   & Kappa & & & & & 95.52 \\\cline{2-8}
                              & \multirow{3}*{Indian Pines} & OA & 90.16 & & 94.34 & & \textbf{97.76} \\
                              &                    & AA & & & & & 98.88 \\
                              &                   & Kappa & & & & & 97.32 \\\cline{2-8}
                              & \multirow{3}*{Pavia University} & OA & 92.56 & & 96.48 & & \textbf{99.31} \\
                              &                    & AA & & & & & 99.17 \\
                              &                   & Kappa & & & & & 99.07 \\\cline{2-8}
    \hline
    \hline
    \multirow{9}*{Setting 2} & \multirow{3}*{KSC} & OA & & 96.31 & & & \textbf{99.09} \\
                              &                    & AA & & 94.68 & & & \textbf{98.26} \\
                              &                   & Kappa & & 95.90 & & & \textbf{98.99} \\\cline{2-8}
                              & \multirow{3}*{Indian Pines} & OA & & 97.56 & & & \textbf{98.01} \\
                              &                    & AA & & \textbf{99.23} & & & 99.07 \\
                              &                   & Kappa & & 97.02 & & & \textbf{97.90} \\\cline{2-8}
                              & \multirow{3}*{Pavia University} & OA & & 99.54 & & & \textbf{99.72} \\
                              &                    & AA & & \textbf{99.66} & & & 99.62 \\
                              &                   & Kappa & & 99.41 & & & \textbf{99.62} \\\cline{2-8}
    \hline
    \hline
    \multirow{9}*{Setting 3} & \multirow{3}*{Pavia Center} & OA & & & & 99.68 & \textbf{99.85} \\
                              &                    & AA & & & & 99.26 & \textbf{99.70} \\
                              &                   & Kappa & & & & 99.55 & \textbf{99.79} \\\cline{2-8}
                              & \multirow{3}*{Indian Pines} & OA & & & & \textbf{99.04} & 98.81 \\
                              &                    & AA & & & & 99.14 & \textbf{99.31} \\
                              &                   & Kappa & & & & \textbf{98.87} & 98.60 \\\cline{2-8}
                              & \multirow{3}*{Pavia University} & OA & & & & 99.08 & \textbf{99.26} \\
                              &                    & AA & & & & 99.08 & \textbf{99.18} \\
                              &                   & Kappa & & & & 98.79 & \textbf{99.03} \\
    \hline
  \end{tabular}
\end{table*}

\begin{table*}[htp]\footnotesize
  \centering
  \caption{NETWORK COMPLEXITIES OF DEEPING LEARNING MODELS ON DEFFIRENT DATA SETS}\label{various_parameters}
  \begin{tabular}{|c|c|c|c|c|c|}
    \hline
    Dataset & CNN & 3D-CNN & CNN-PPF & S-CNN & FSSF-Net \\
    \hline
    Salinas & 82,216 & & 178,278 & & 123,696 \\\hline
    Indian Pines & 81,408 & 45,331,024 & 65,019 & 3,331,900 & 125,296 \\\hline
    Pavia University & 61,249 & 5,860,841 & 33,019 & 2,207,480 & 77,854 \\\hline
    Kennedy Space Center  & & 5,987,437 & & & 105,578 \\\hline
    Pavia Center & & & & 2,206,940 & 77,754 \\
    \hline
  \end{tabular}
\end{table*}

The next experiment is to evaluate the HSI classification methods in the situation of small size of training samples.
As we know, the deep learning technique always need a significant amount of labeled data for training.
However, due to the limit of available labeled data in HSI, it is necessary to evaluate the deep learning methods using limited training samples.
Four spectral-spatial deep learning models 3D-CNN, CNN-PPF, S-CNN, and FSSF-Net are evaluated,
and we also choose two prominent traditional methods, SVM, and 3D-DWT, so as to contrast to these deep learning methods.
In our experiments, we randomly select 50 samples from each class to form our training set, leaving the rest as the test set.
Salinas, Indian Pines, and Pavia University are used as our experimental data sets.
Especially, as for Indian Pines, because some classes have fewer than 50 labeled samples, we only choose 9 classes that contain more than 400 samples for classification.
The experimental setting of our method maintains the same as stated in Section~\ref{sec:framework_setting}, and for the other methods their settings follows their original papers.

The compared results are listed in Tables~\ref{indian_comp_results},~\ref{salinas_comp_results},~\ref{paviaU_comp_results}.
Compared with the classification results in Table \ref{various_settings}, all deep learning methods suffer the performance degradation in different degrees.
However, our method consistently outperforms the other ones,
indicating that our method can better deal with small training sample size problem by taking full advantage of spectral and spatial properties of HSI in a well-designed network architecture.
As the size of the training set decreases, there is more impact on 3D-CNN and S-CNN than CNN-PPF, because CNN-PPF has fewer parameters and is less likely to over-fitting.  Surprisingly, 3D-DWT achieves pretty competitive results when comparing to deep learning methods.

We further drill down to details of these deep learning methods and compare to our model.
To overcome the problem of small size of available labeled samples in HSI,
CNN-PPF and S-CNN use data augmentation technique, in which each pixel is combined with other pixels belonging to the same or different classes to form the paired training samples. In our model, we merge spectral and spatial properties of HSI into the network structure, so it also plays a role of data augmentation but by a new way in which the central pixel in a patch is enhance by the neighboring pixels with or without lables.
However, both CNN-PPF and S-CNN are just  feature extractors and adopt separated classifiers to classify the extracted features.
On the contrary, our model is end-to-end, i.e., combines the feature extraction with classification at the same time, which is not only easy to implement, but improve the performance of feature extraction and classification as well.
3D-CNN also takes a patch in HSI as its input the same as our method, but it has not the sharing architecture used in our method, so it contains more parameters, making it work worse than our model in the case of small training sample size.

\begin{figure}[!htp]
  \centering
  \includegraphics[width=0.6\linewidth]{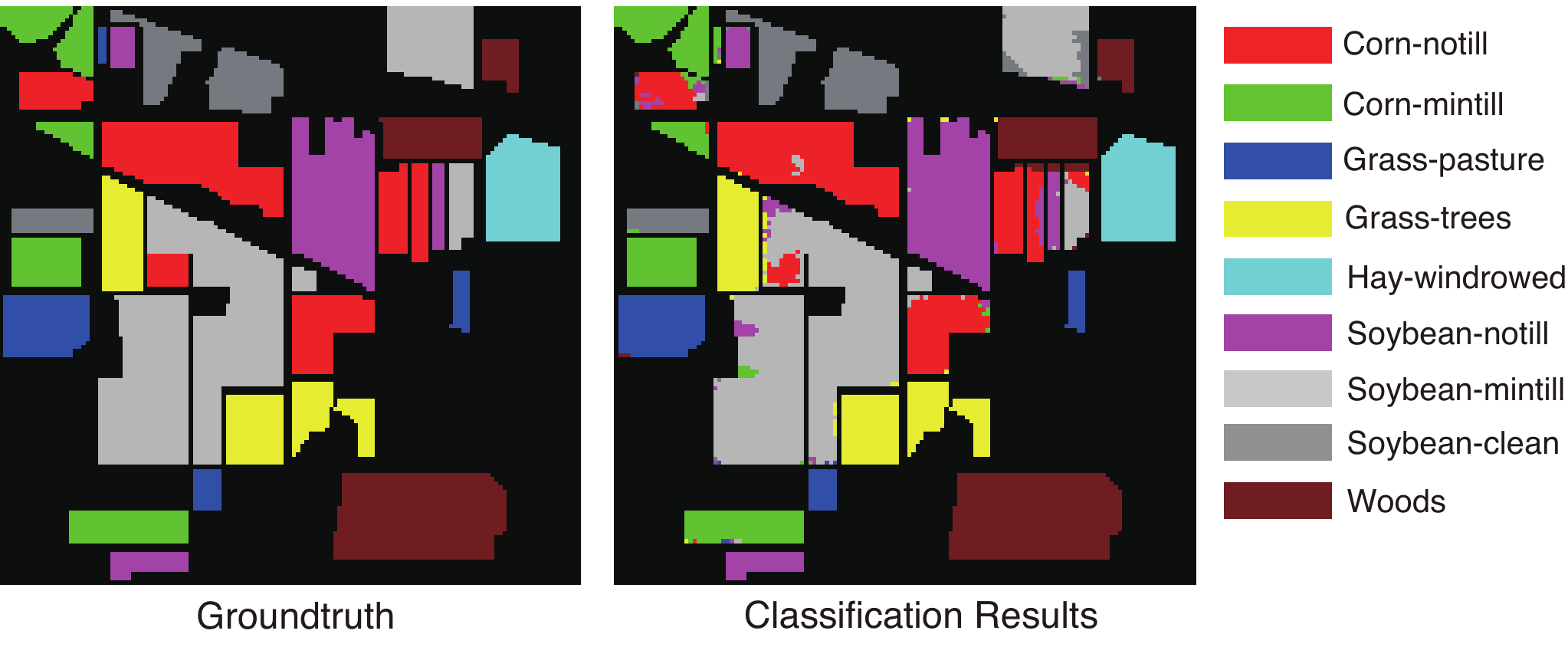}
  \caption{Classification Map of Indian Pines}\label{fig:cls_maps_IN}
\end{figure}

\begin{figure}[!htp]
  \centering
  \includegraphics[width=0.6\linewidth]{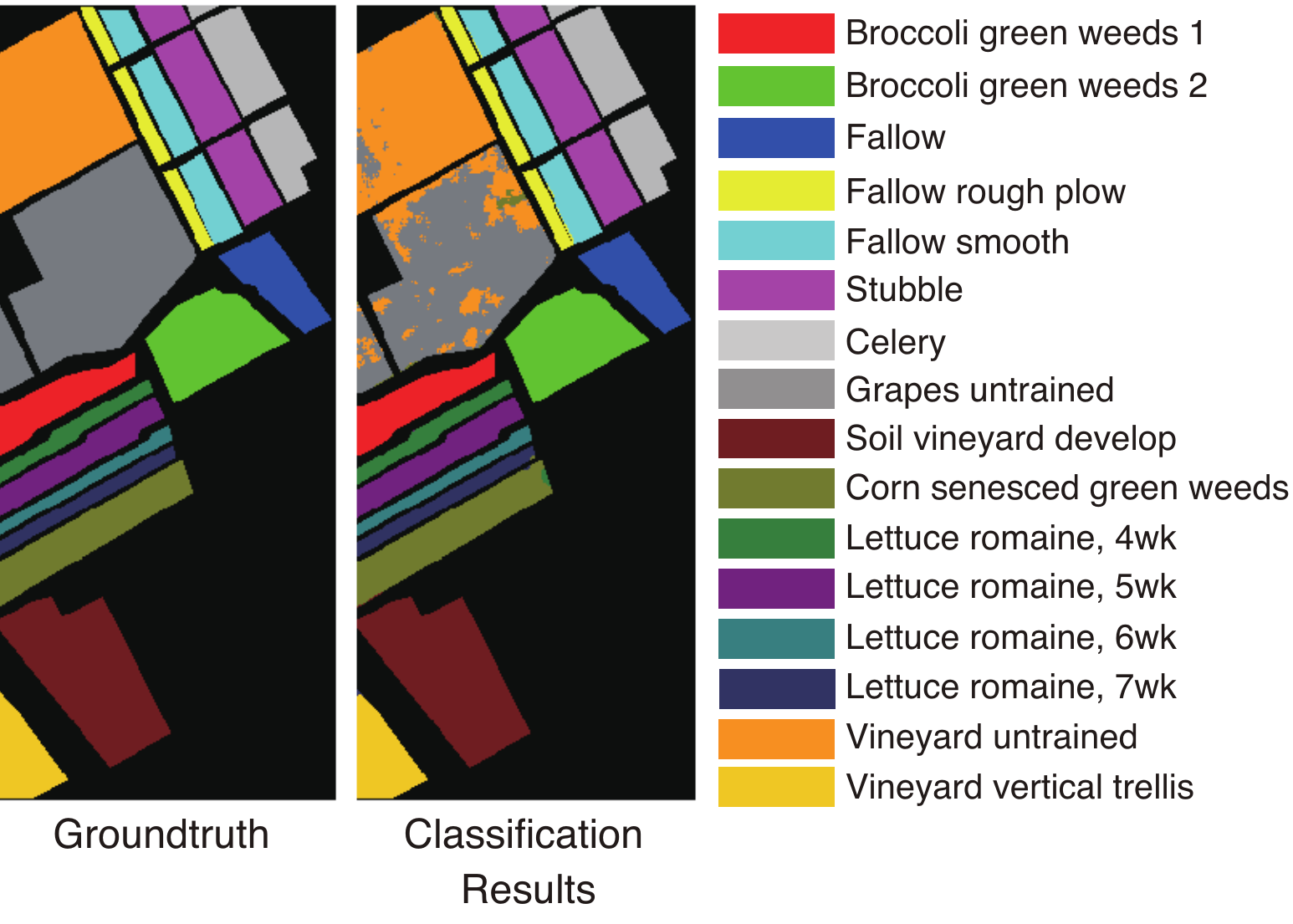}
  \caption{Classification Map of Salinas}\label{fig:cls_maps_SA}
\end{figure}

\begin{figure}[!htp]
  \centering
  \includegraphics[width=0.6\linewidth]{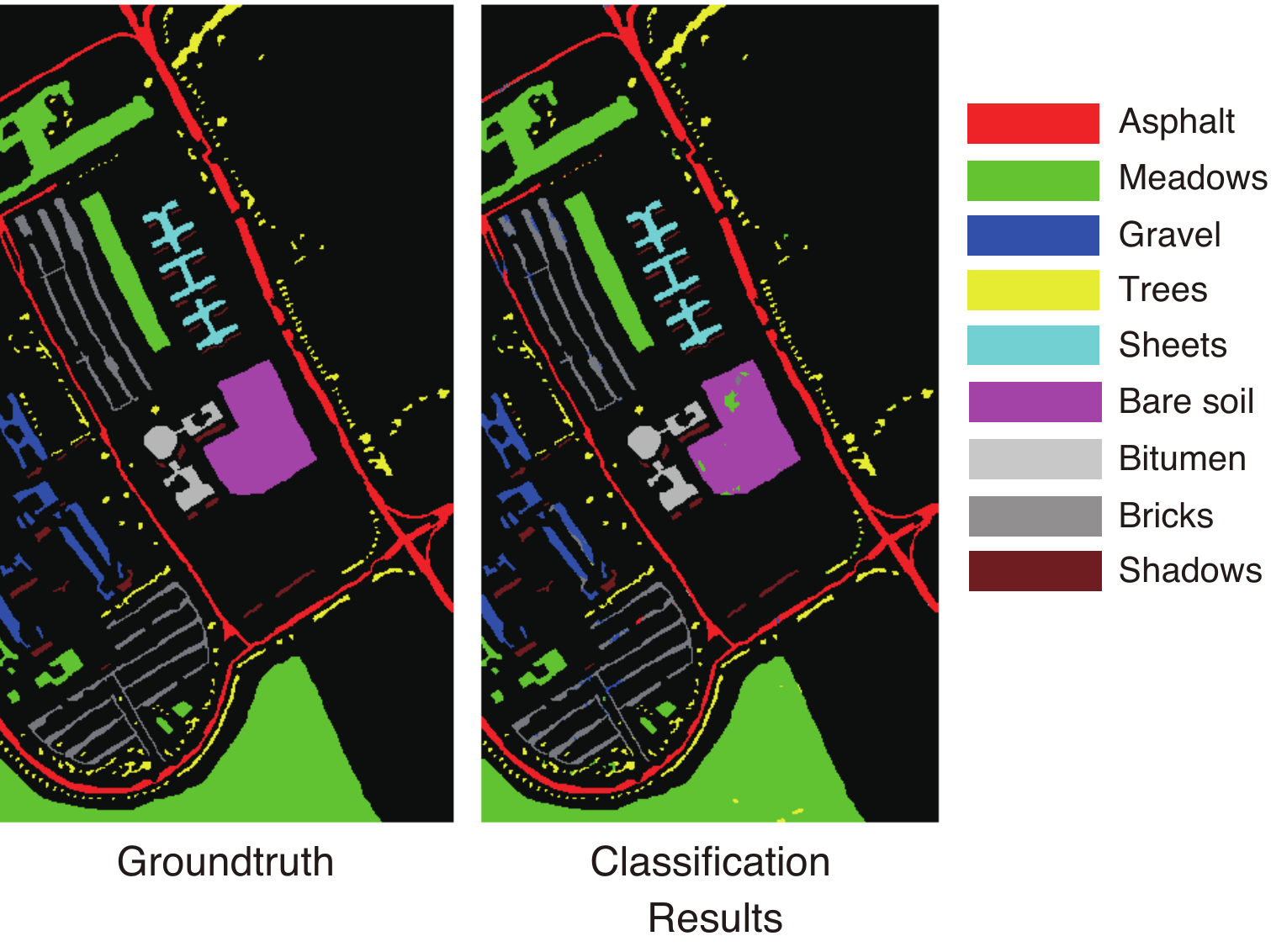}
  \caption{Classification Map of Pavia University}\label{fig:cls_maps_PU}
\end{figure}

As a supplement, we further record the training and inference time in Table~\ref{execu_time}, from which
it can be found that our method converges faster than 3D-CNN, CNN-PPF, and S-CNN, but costs more time than CNN-PPF and S-CNN to inference.
The reason may be that in our method, the pre-training PSE-Net $g$ needs to process each pixel in a patch individually whereas CNN-PPF and S-CNN process the patch as a whole once.
Moreover, for better visualization of classification performance, we plot classification maps of our method on three data sets, see Figs.~\ref{fig:cls_maps_IN},~\ref{fig:cls_maps_SA},~\ref{fig:cls_maps_PU}. Comparing to the corresponding ground truth, our results are pretty close to the real ones except failing at some verges of certain patches in the classification maps.

\begin{table*}[!htp]\footnotesize
  \centering
  \caption{Classification results (\%) OF with 50 training SAMPLES OF EACH CLASS in INDIAN PINES}\label{indian_comp_results}
  \begin{tabular}{|c|c|c|c|c|c|c|}
    \hline
    \multirow{2}*{Class} & \multirow{2}*{SVM} & \multirow{2}*{3D-DWT} & \multirow{2}*{3D-CNN} & \multirow{2}*{CNN-PPF} & \multirow{2}*{S-CNN} & \multirow{2}*{FSSF-Net} \\
      &   &   &   &   &   &   \\
    \hline
    \hline
    1 & 48.72 & 79.74 & 77.39 & 80.19 & 18.07 & 92.45 \\
    2 & 45.74 & 86.72 & 85.00 & 91.92 & 42.82 & 98.74 \\
    3 & 81.20 & 97.41 & 92.98 & 97.23 & 86.14 & 98.80 \\
    4 & 95.18 & 96.82 & 96.53 & 99.56 & 97.21 & 99.65 \\
    5 & 96.26 & 99.72 & 99.07 & 99.77 & 97.66 & 100  \\
    6 & 58.35 & 83.95 & 83.60 & 88.07 & 24.62 & 92.23 \\
    7 & 52.67 & 77.09 & 67.05 & 76.72 & 53.35 & 85.61 \\
    8 & 59.04 & 88.73 & 90.53 & 92.82 & 67.96 & 95.10 \\
    9 & 87.59 & 98.57 & 94.81 & 99.67 & 93.58 & 99.59 \\
    \hline
    \hline
    OA & 64.08 & 86.40 & 82.42 & 87.88 & 57.50 & \textbf{93.50} \\
    \hline
    AA & 69.42 & 89.86 & 87.44 & 91.77 & 64.60 & \textbf{95.80} \\
    \hline
    Kappa & 58.33 & 84.10 & 79.59 & 85.84 & 51.09 & \textbf{92.39} \\
    \hline
  \end{tabular}
\end{table*}

\begin{table*}[!htp]\footnotesize
  \centering
  \caption{Classification results (\%) with 50 training SAMPLES OF EACH CLASS in SALINAS}\label{salinas_comp_results}
  \begin{tabular}{|c|c|c|c|c|c|c|}
    \hline
    \multirow{2}*{Class} & \multirow{2}*{SVM} & \multirow{2}*{3D-DWT} & \multirow{2}*{3D-CNN} & \multirow{2}*{CNN-PPF} & \multirow{2}*{S-CNN} & \multirow{2}*{FSSF-Net} \\
      &   &   &   &   &   &   \\
    \hline
    \hline
    1 & 98.82 & 98.39 & 98.42 & 99.80 & 99.95 & 100 \\
    2 & 99.31 & 97.67 & 92.86 & 99.54 & 53.56 & 100 \\
    3 & 97.90 & 97.93 & 97.59 & 99.79 & 59.71 & 100 \\
    4 & 99.38 & 98.78 & 99.87 & 99.85 & 100 & 99.97 \\
    5 & 97.05 & 98.93 & 99.29 & 95.51 & 99.39 & 99.39 \\
    6 & 99.52 & 99.37 & 99.22 & 99.69 & 99.87 & 100 \\
    7 & 99.42 & 98.16 & 94.66 & 99.77 & 97.00 & 99.99 \\
    8 & 64.11 & 79.29 & 69.95 & 89.15 & 74.32 & 81.23 \\
    9 & 98.69 & 97.71 & 95.37 & 98.68 & 96.86 & 100  \\
    10 & 89.23 & 92.43 & 97.68 & 93.25 & 79.18 & 97.22 \\
    11 & 97.31 & 99.61 & 98.45 & 99.31 & 96.66 & 99.69 \\
    12 & 99.40 & 99.98 & 98.87 & 100 & 100 & 100  \\
    13 & 97.64 & 97.74 & 99.31 & 99.31 & 99.88 & 99.95 \\
    14 & 94.59 & 96.08 & 98.86 & 96.67 & 96.08 & 99.76 \\
    15 & 71.17 & 81.17 & 84.13 & 68.36 & 73.75 & 87.71 \\
    16 & 98.43 & 98.45 & 96.98 & 98.98 & 97.61 & 99.44 \\
    \hline
    \hline
    OA & 87.08 & 91.65 & 89.57 & 92.33 & 84.30 & \textbf{94.16} \\
    \hline
    AA & 93.87 & 95.73 & 95.09 & 96.10 & 88.99 & \textbf{97.77} \\
    \hline
    Kappa & 85.65 & 90.72 & 88.45 & 91.57 & 82.56 & \textbf{93.50} \\
    \hline
  \end{tabular}
\end{table*}

\begin{table*}[!htp]\footnotesize
  \centering
  \caption{Classification results (\%) with 50 training SAMPLES OF EACH CLASS in PAVIA UNIVERSITY }\label{paviaU_comp_results}
  \begin{tabular}{|c|c|c|c|c|c|c|}
    \hline
    \multirow{2}*{Class} & \multirow{2}*{SVM} & \multirow{2}*{3D-DWT} & \multirow{2}*{3D-CNN} & \multirow{2}*{CNN-PPF} & \multirow{2}*{S-CNN} & \multirow{2}*{FSSF-Net} \\
      &   &   &   &   &   &   \\
    \hline
    \hline
    1 & 75.50 & 91.72 & 84.00 & 95.40 & 86.22 & 95.22 \\
    2 & 79.88 & 93.74 & 94.23 & 87.08 & 95.37 & 97.14 \\
    3 & 77.99 & 84.56 & 84.19 & 90.24 & 76.28 & 90.39 \\
    4 & 92.49 & 95.57 & 97.31 & 92.80 & 97.84 & 95.98 \\
    5 & 99.55 & 99.60 & 99.74 & 99.92 & 100 & 100 \\
    6 & 80.37 & 90.83 & 88.93 & 92.70 & 82.95 & 93.83 \\
    7 & 91.84 & 96.59 & 92.92 & 94.38 & 88.59 & 100 \\
    8 & 79.53 & 91.87 & 91.21 & 87.94 & 77.73 & 95.79 \\
    9 & 99.60 & 99.89 & 97.61 & 99.22 & 99.55 & 99.84 \\
    \hline
    \hline
    OA & 81.42 & 93.01 & 91.69 & 90.54 & 90.25 & \textbf{96.16} \\
    \hline
    AA & 86.31 & 93.82 & 92.24 & 93.30 & 89.39 & \textbf{96.47} \\
    \hline
    Kappa & 76.10 & 90.79 & 89.06 & 87.67 & 87.06 & \textbf{94.91} \\
    \hline
  \end{tabular}
\end{table*}

\begin{table}[!htp]\footnotesize
  \centering
  \caption{EXECUTION TIME (SECOND) OF TRAINING AND TESTING PROCEDURES}\label{execu_time}
  \begin{tabular}{|c|c|c|c|c|}
    \hline
    \multirow{2}*{Methods} & \multirow{2}*{Procedure} & Univeristy of & Indian & \multirow{2}*{Salinas} \\
                           &  & Pavia & Pines & \\
    \hline
    \hline
    \multirow{2}*{3D-CNN} & training & 1498 & 5387 & 9130 \\
                       & testing & 106 & 82 & 568 \\
    \hline
    \multirow{2}*{CNN-PPF} & training  & 726 & 952 & 2431 \\
                           & testing  & 14 & 4 & 18 \\
    \hline
    \multirow{2}*{S-CNN} & training  & 120 & 558 & 795 \\
                         & testing  & 2 & 1 & 2 \\
    \hline
    \multirow{2}*{FSSF-Net} & training & 357 & 370 & 401 \\
                              & testing & 65 & 17 & 84 \\
    \hline
  \end{tabular}
\end{table}

\section{Conclusions}\label{sec:conclusion}

Deep learning applied for the HSI classification is a fast developing area.
Motivated by the spectral-spatial structure of HSIs, a novel end-to-end network architecture is proposed in this paper, which factorizes spectral-spatial feature into two subnetworks. Because of rich spectral information of the hyperspectral data, pretraining the subnetwork used for spectral feature extracting with labeled samples not only reduces the feature dimensionality but also learns inherent spectral structure that is better for classification.
To fuse the local spatial information, we share the pre-trained subnetwork in a patch and combine the extracted features into a vector feeding to another subnetwork for final classification. The sharing scheme can decrease the number of parameters to avoid over-fitting problem.
Compared with some state of the art deep learning methods and  typical conventional classification methods, the proposed method consistently outperforms them in all  experiments.

\newpage
{\small
\bibliographystyle{ieee}
\bibliography{egpaper_final}
}

\end{document}